\documentclass[graybox]{svmult}

\usepackage{type1cm}        
\usepackage{makeidx}         
\usepackage{graphicx}        
\usepackage{orcidlink}

\usepackage{multicol}        
\usepackage[bottom]{footmisc}

\usepackage{newtxtext}       %
\usepackage{newtxmath}       
\usepackage{hyperref}
\usepackage{nameref}

\makeindex             

\begin{document}

\title*{A Fuzzy Evaluation of Sentence Encoders on Grooming Risk Classification}
\author{Geetanjali Bihani\orcidlink{0000-0001-8352-7948}, Julia Rayz\orcidlink{0000-0003-3786-2416}}
\institute{Geetanjali Bihani \at Purdue University, USA, \email{gbihani@purdue.edu},
\and Julia Rayz \at  Purdue University, USA,  \email{jtaylor1@purdue.edu}}
%
%
\maketitle

\abstract*{}

\abstract{
With the advent of social media, children are becoming increasingly vulnerable to the risk of grooming in online settings. Detecting grooming instances in an online conversation poses a significant challenge as the interactions are not necessarily sexually explicit, since the predators take time to build trust and a relationship with their victim. Moreover, predators evade detection using indirect and coded language. While previous studies have fine-tuned Transformers to automatically identify grooming in chat conversations, they overlook the impact of coded and indirect language on model predictions, and how these align with human perceptions of grooming.
In this paper, we address this gap and evaluate bi-encoders on the task of classifying different degrees of grooming risk in chat contexts, for three different participant groups, i.e. law enforcement officers, real victims, and decoys. Using a fuzzy-theoretic framework presented in \cite{bihanirayz24_regress}, we map human assessments of grooming behaviors to estimate the actual degree of grooming risk. Our analysis reveals that fine-tuned models fail to tag instances where the predator uses indirect speech pathways and coded language to evade detection. Further, we find that such instances are characterized by a higher presence of out-of-vocabulary (OOV) words in samples, causing the model to misclassify. Our findings highlight the need for more robust models to identify coded language from noisy chat inputs in grooming contexts.} 

\section{Introduction}
\label{sec:1}
Grooming refers to the insidious process where predators forge a relationship with children and build up trust with the intention of sexual exploitation \cite{ring21}.  In \cite{bihanirayz24_regress}, we noted that models assign much lower risk scores to high-risk chat contexts because these contexts contain implicit and coded language. In this work, we follow up on this initial investigation and evaluate Transformer-based classifiers on the task of detecting different degrees of grooming risk in online grooming conversations. We approach this task with a non-binary perspective, recognizing the necessity to differentiate between different levels of risk, because it can inform targeted preventive measures to safeguard individuals in realistic scenarios.

Grooming has been described as a complex multi-stage phenomenon, with the severity of grooming varying as the chat progresses\cite{ring21, olson2007}. Yet, works on automated grooming detection usually frame it as a binary classification problem \cite{preub2021, vogt2021}, categorizing entire chat instances as either grooming or non-grooming. Existing methods for grooming detection primarily rely on decoy conversations to train and fine-tune neural language models \cite{vogt2021}. Whether decoy conversations can be used to approximate real grooming scenarios has been contested in linguistic and cyberforensic analyses of online grooming conversations \cite{chiang19, ring21_2, ring22}. This disparity in data distribution can have an impact on the generalization of such models when applied out of distribution (OOD). 

We aim to address these gaps in this paper by evaluating Transformer sentence encoders on the task of classifying degrees of grooming risk expressed in natural language. We evaluate these models across diverse participant groups, specifically for real victims, law enforcement officers (LEO), and decoys. To that end, we use the fuzzy-theoretic framework that maps human assessments of grooming behaviors present within chats to the severity of grooming risk, as given in \cite{bihanirayz24_regress}. Our analysis reveals that fine-tuned sentence encoder classifiers show an increased rate of errors in identifying high-risk chat contexts, which is caused by the indirect speech pathways used by predators to manipulate and coerce victims. We find that Transformer classifiers fail to flag cases that contain coded language and lack sexually explicit content. Further, we find that the proportion of coded language used in grooming chats across different participant groups varies, causing the models to have different performance across populations. 

This finding underscores the importance of robust modeling of indirect speech acts by language models, especially those utilized by predators. Notably, to the best of our knowledge, no prior work has incorporated human assessments of risk to compare fine-tuning results, making our study unique in its comprehensive evaluation of language model performance in grooming risk classification.




\section{Categorizing Grooming Risk}
\label{sec:2}

The concept of risk, much like other subjective notions, embodies fuzziness. This fuzziness becomes apparent when considering grooming strategies, where what constitutes risky behavior and what does not is often blurry. These degrees of severity are influenced by the presence of grooming strategies within a given context \cite{ring21}. Moreover, grooming strategies employed by individuals with malicious intent, can vary widely in their subtlety and perceived harm \cite{ring21}. Similar to the concept of risk, the assessment of grooming strategies itself lacks a precise boundary, as it depends on various factors such as individual perceptions, cultural norms, and situational contexts.

\begin{figure}
\centering
\includegraphics[scale=.2]{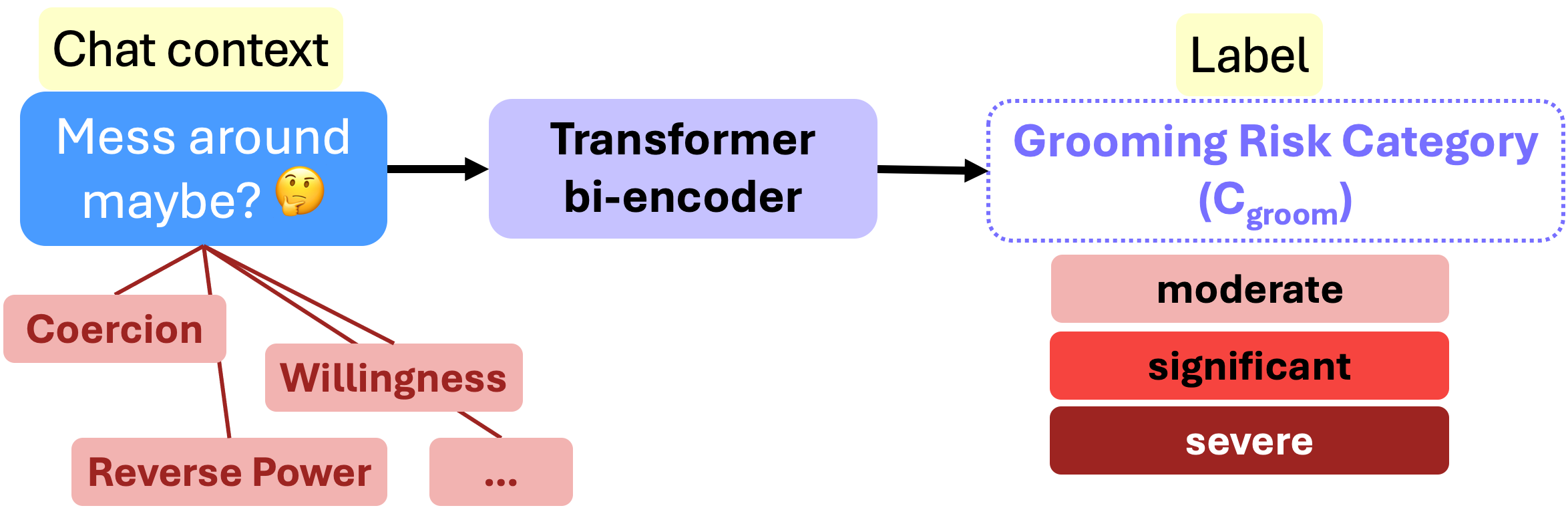}
\caption{Presence of grooming strategies present within a given chat context used to create grooming risk categories}
\label{fig:pipeline}       
\end{figure}


In this work, we use grooming strategies present within a given text to create grooming risk categories using membership functions defined in \cite{bihanirayz24_regress}. Cyberforensic analyses of grooming conversations have shown a higher presence of grooming strategies within more advanced stages of grooming \cite{ring21}. Based on these findings, we utilize human annotations for the presence of twelve grooming strategies, as described in \cite{ring21} to estimate the degree of risk in a chat context. For a given chat context $c$, we define the total number of observed grooming strategies $o(c)$ as the sum of individual strategy scores $s_i$, where each $s_i \in \{0, 0.5, 1\}$ represents the absence (0), partial presence (0.5), or full presence (1) of the $i^{th}$ strategy. This is shown in Equation~\ref{eq:1}, also described in prior work \cite{bihanirayz24_regress}. Ideally, the maximum number of strategies assigned to a single chat line can be 12. However, we find that in our data, such cases do not occur, with the highest number of strategies used together limited to 5. This can be attributed to the limited context that the predator has to work with, where they test out different strategies during different stages of the conversation \cite{kloess2017}. For a detailed description of the grooming strategies used in this work, please refer to Table~\ref{app_tab:1} in the Appendix.

\begin{equation}
\label{eq:1}
o(c)=\sum_{i=1}^{N} s_{i}(c)
\end{equation}

%

\begin{equation}
\label{eq:2}
\mu_{risk}(c) = \varphi\left(o(c)-m\right) \text{, where }\varphi(z)=\frac{e^{-z^{2} / 2}}{\sqrt{2 \pi}}
\end{equation}
\begin{equation}
\label{eq:2.1}
\mu_{mod}(c) = \mu^{0.2}_{risk}(c) \text{ for $m=0.2$}
\end{equation}
\begin{equation}
\label{eq:2.2}
\mu_{sig}(c) = \mu_{risk}(c) \text{ for $m=1$}
\end{equation}
\begin{equation}
\label{eq:2.3}
\mu_{sev}(c) = \mu^2_{risk}(c) \text{ for $m=2$}
\end{equation}

We employ a Gaussian membership function to map the overall strategy membership $(o(c))$ to various degrees of grooming risk, as outlined in Equation~\ref{eq:2}. While this equation was originally introduced in \cite{bihanirayz24_regress} for fuzzy evaluation of grooming risk scoring, we leverage this framework to delineate grooming risk into three distinct severity levels, which are not necessarily mutually exclusive. These categories are defined as \textit{moderate}, \textit{significant}, and \textit{severe} levels of grooming risk, arranged in ascending order of risk severity. The respective fuzzy membership functions for moderate, significant, and severe risk are defined in Equations~\ref{eq:2.1}, \ref{eq:2.2} and \ref{eq:2.3}. Using this method, chat contexts with fewer grooming strategies are assigned higher membership values within the \textit{moderate} risk category. Conversely, those with a greater number of strategies present receive higher memberships within the \textit{significant} and \textit{severe} grooming risk categories. For our classifier fine-tuning process, we defuzzify the risk degree using an $\alpha$-cut$=0.5$, selecting the highest degree of risk membership exceeding the $\alpha$-cut. Thus, if a chat context has memberships surpassing the $\alpha$-cut in \textit{moderate} and \textit{severe} categories, it is categorized as a \textit{severe} risk chat context.




\section{Methodology}
\label{sec:3}
This section details our approach to fine-tuning bi-encoders specifically for grooming risk classification, utilizing human assessments of grooming strategy behaviors in chat contexts. Our aim in this investigation is to evaluate the potential of fine-tuning models from the Transformer bi-encoder family \cite{sbert2019}, to estimate the extent of grooming risk present in chat interactions. Our focus extends beyond conventional language cues, as we aim to discern the limitations of these bi-encoders, particularly in scenarios where higher-risk situations may not always manifest through explicit language. Drawing from prior research indicating variations in language usage across different groups in grooming behaviors \cite{ring21}, we fine-tuned and evaluated our bi-encoders separately for distinct grooming contexts. Specifically, we focused on grooming conversations involving predators interacting with law enforcement officers (LEO), victims, and decoys. This analysis underscores the oversight of current automated models of grooming classification in accounting for nuanced differences across different participant conversations. 

\subsection{Task Definition}
\label{subsec:3.1}

\textbf{Definition 1 - Chat window length $(w)$}: Number of messages in an exchange between two participants in a grooming conversation.

\noindent \textbf{Definition 2 - Chat Context $(c)$}: A sequence comprising the current and last $w-1$ messages in a grooming conversation. We fix $w=4$ for our analysis.

\noindent \textbf{Definition 3 - Grooming risk category $(C_{groom})$}: Denotes the severity of grooming within a chat context within the following three categories: \{\textit{moderate}, \textit{significant}, \textit{severe}\}.

\subsection{Models Studied}
\label{subsec:2}
We conducted our analysis on pre-trained bi-encoder models. These encoders leverage siamese and triplet network structures to derive sentence embeddings with semantically meaningful representations. These representations are optimized to capture semantic similarity between sentences in a vector space, making them suitable for various downstream tasks such as semantic search and clustering. The models we analyze include Sentence-BERT (SBERT) \cite{sbert2019}, MPNET \cite{mpnet}, and RoBERTa \cite{roberta}. The base models, as given within the names, are BERT, MPNET and RoBERTa respectively. We choose BERT and RoBERTa based on their application in prior work, and MPNET based on the quality of embeddings it generates.

\subsection{Fine-tuning Details}
\label{subsec:3}
The process of fine-tuning involves adapting pre-trained models to specific downstream tasks using task-specific data. We fine-tune bi-encoders to predict grooming risk class for a given chat context $c$ by optimizing on minimizing the cross-entropy loss between the predicted and actual grooming risk classes. For model fine-tuning, we use an Adam optimizer \cite{kingma2014}, learning rate of $2.10^{-5}$, over $5$ epochs, with a batch size of $4$.

To understand the performance differences of models fine-tuned on interactions of predators with different participant groups, we fine-tuned three separate models. Thus, one model underwent fine-tuning on interactions of predators with law enforcement officers (LEO), another on interactions with real victims (Victim), and a third on interactions with decoys (Decoy).

\section{Results}
We examined how well fine-tuned bi-encoder predictions fare on the task of identifying varying degrees of grooming risk in predatory chat contexts, across different participant groups. We consider chat contexts with a window size of $w=4$. Our analysis reveals variations in model performance across different levels of grooming risk, with the model achieving higher accuracy in \textit{moderate} contexts but showing poorer performance in \textit{significant} and \textit{severe} risk scenarios. 




We report macro-averaged F1 scores on predictions in Table~\ref{tab:3}, and note a stark difference between F1 scores on \textit{moderate}, \textit{significant} and \textit{severe} risk contexts. We find that across the three participant groups, \textit{moderate} and \textit{severe} risk contexts receive higher coverage as compared to \textit{significant} risk contexts. These results highlight the limitations of fine-tuned bi-encoder models in detecting grooming behaviors where human evaluators have indicated a higher presence of grooming strategies. Moreover, all three models show similar error rates across different degrees of risk. These findings can be attributed to their inability to capture indirect communication pathways and adversarial grooming entrapment language (e.g. misspelled words, abbreviations, emojis, etc.) utilized by predators \cite{Lykousas2021}. We also find that the models show the worst performance for \textit{significant} risk scenarios for decoy chats while showing a ~$10\%$-$20\%$ better F1 score for law enforcement and real victim chats. This can be attributed to the differences in the way grooming conversations progress for real victims versus decoys \cite{chiang19}. These findings highlight the differences in grooming detection model performance across different participant groups and question the inherent assumption made by the prior work regarding training and finetuning automated models of grooming risk estimation using unrepresentative decoy and law enforcement chats.

\begin{table}[!t]
\centering
\caption{F1 score reported for different participant groups}
\label{tab:3}       
%
%
\begin{tabular}{p{0.2\textwidth}p{0.2\textwidth}p{0.1\textwidth}p{0.1\textwidth}p{0.1\textwidth}}
\hline\noalign{\smallskip}
\textbf{Model} & \textbf{Grooming Risk}  & \textbf{LEO}  & \textbf{Victim}  & \textbf{Decoy}  \\
\noalign{\smallskip}\hline\noalign{\smallskip}
$S_{\text {\textbf{\textit{BERT-base}}}}$ & Moderate &  0.83 & 0.90 & 0.84 \\
& Significant & 0.47 & 0.39 & 0.26 \\
& Severe & 0.53 & 0.74 &  0.67 \\
\noalign{\smallskip}\hline\noalign{\smallskip}
& Overall & 0.70 & 0.68 & 0.59  \\
\noalign{\smallskip}\hline\noalign{\smallskip}

$S_{\text {\textbf{\textit{RoBERTa-base}}}}$ & Moderate & 0.84  & 0.89 & 0.83 \\
& Significant & 0.49 & 0.43 & 0.27 \\
& Severe & 0.80 & 0.68 & 0.71 \\
\noalign{\smallskip}\hline\noalign{\smallskip}
& Overall & 0.71 & 0.67 & 0.60 \\
\noalign{\smallskip}\hline\noalign{\smallskip}

$S_{\text {\textbf{\textit{MPNet}}}}$ & Moderate & 0.83 & 0.90 & 0.86 \\
& Significant & 0.46 & 0.42 & 0.27 \\
& Severe & 0.82 & 0.71 & 0.67 \\
\noalign{\smallskip}\hline\noalign{\smallskip}
& Overall & 0.71 & 0.68 & 0.60  \\
\noalign{\smallskip}\hline\noalign{\smallskip}
\end{tabular}
\end{table}

%
\begin{figure}[htbp]
\centering
\includegraphics[scale=.12]{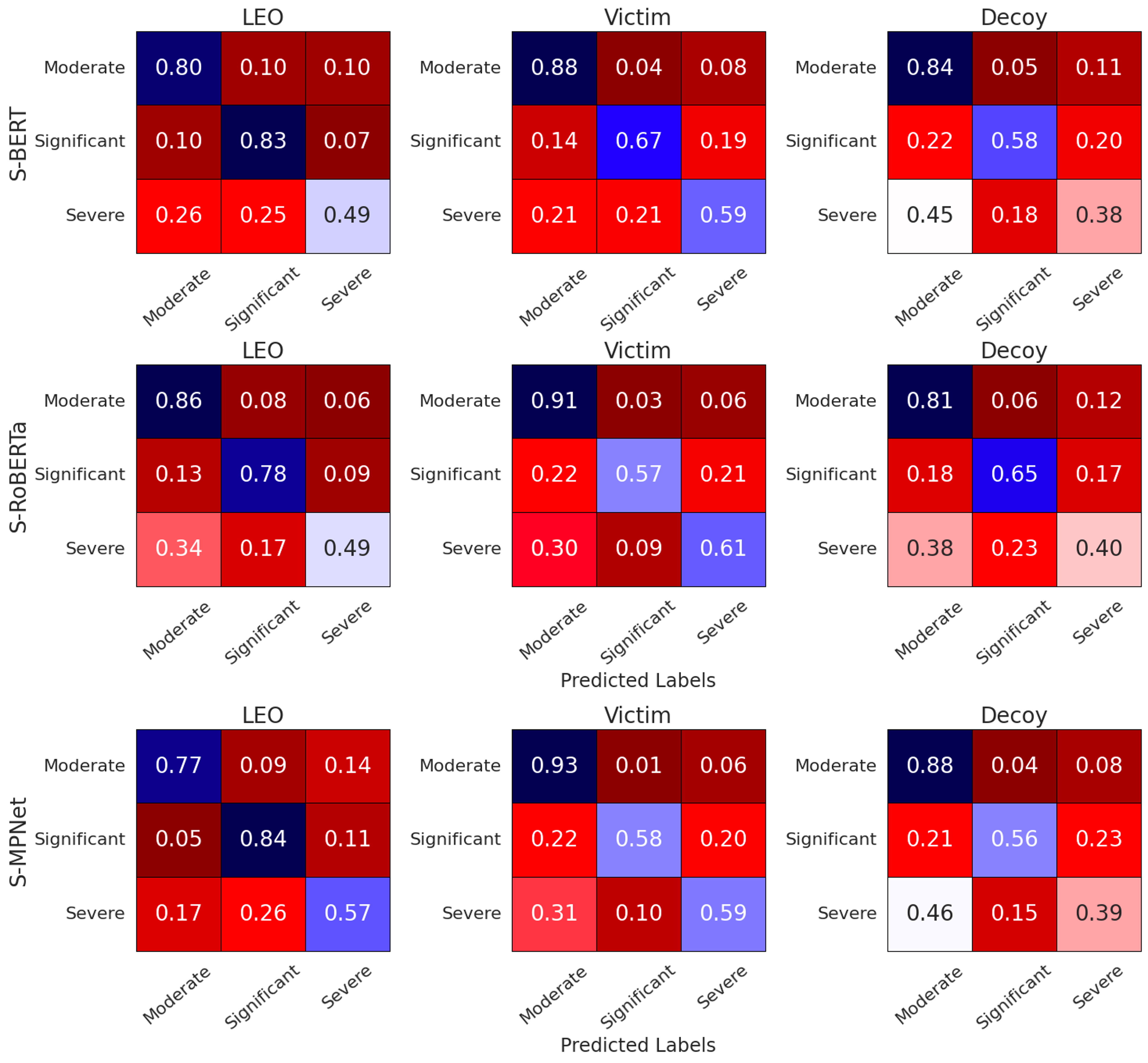}
\caption{Confusion matrices illustrating the classification performance of the models across different participant groups: law enforcement officers (LEO), victims, and decoys. Each heatmap represents the comparison between actual and predicted labels for grooming risk categories (moderate, significant, severe). The top label refers to the participant group on which the model was fine-tuned. The color intensity denotes the count of instances falling into each category, with blue shades indicating higher counts, and red shades indicating lower counts. }
\label{fig:heatmap}       
\end{figure}

We also analyzed model predictions across different participant groups, reporting the classification performance as depicted in Figure~\ref{fig:heatmap}. We find that for law enforcement conversations, the model's classification performance has high coverage across all severities of grooming risk. On the other hand, the model incorrectly classifies more than~$40\%$ of \textit{significant} and \textit{severe} risk samples for decoy chats. Similar to the findings in \cite{bihanirayz24_regress}, our model is unable to detect \textit{significant} and \textit{severe} grooming risk contexts due to an absence of explicit sexual and predatory language in many cases. Instead, predators adopt the use of adversarial grooming entrapment language (e.g. misspelled words, abbreviations, emojis, etc.), which lends to a higher occurrence of out-of-vocabulary (OOV) tokens in the fine-tuning set. Refer to Table ~\ref{tab:oov} for a comparison of the presence of OOV tokens across different participant groups. An increased proportion of OOV tokens in chat instances leads to the deterioration of model performance, causing the model to not flag severe grooming risk instances.  

\begin{table}[htbp]
\caption{OOV tokens across chats from different participant groups}
\label{tab:oov}       

\begin{tabular}{cccc}
\hline\noalign{\smallskip}
 & \textbf{LEO} & \textbf{Victim} & \textbf{Decoy} \\
\noalign{\smallskip}\hline\noalign{\smallskip}
OOV \% & $26.70$ & $20.68$ & $30.80$ \\
OOV words per message chunk & $4.85$ & $4.88$ & $6.48$ \\
\noalign{\smallskip}\hline\noalign{\smallskip}
\end{tabular}
\end{table}

\section{Discussion and Conclusion}
    This paper investigates whether bi-encoders can learn to classify varying degrees of grooming risk inherent in online grooming conversations. We fine-tune and evaluate a transformer bi-encoder on the task of grooming detection across different participant groups, for increasing degrees of risk. Our analysis highlights that fine-tuned sentence encoders still exhibit higher Detecting coded language in harmful contexts is a non-trivial task, particularly in domains like online grooming, where automated models of detection cannot discount precision.  While Transformer bi-encoders like SBERT show promise, they rely on surface-level features and are fine-tuned on decoy conversations instead of real victim chats, leaving questions about their effectiveness in realistic scenarios unanswered. Additionally, previous work has not compared model fine-tuning results with human assessments of grooming risk. This study addresses this gap by investigating the ability of bi-encoders to predict the level of grooming risk in chat contexts, for three different participant groups, i.e. law enforcement officers, real victims, and decoys. in categorizing more severe risk contexts, misclassifying them as the lowest severity defined in this work. We find that such discrepancies are tied to cases where surface form text does not contain explicit identifiers of grooming, but rather uses indirect speech pathways to manipulate victims. Our results align with recent research which shows that many predatory behaviors can evade detection from current filters through tactics like modifying explicit identifiers, introducing typos, using emojis, and out-of-vocabulary words \cite{Lykousas2021}. In such cases, fine-tuning sentence embedding models does not help the model learn useful indicators of grooming within natural language and is a trivial approach to solving a non-trivial task. The sole reliance on word form and incentivizing training loss lead to learning shortcuts while ignoring nuance \cite{bihanirayz2024}.  Even with the integration of long-range context, the task of encoding intricate lexical semantic phenomena to enhance natural language understanding continues to be a challenge \cite{bihanirayz21, vulic2021}. This finding underscores how the current fine-tuning methods for transformer bi-encoders still need to be improved, to be deployed in real-world scenarios and calls for the need for robust modeling of indirect speech acts employed in grooming contexts by language models. We plan to address these findings in future work.

\begin{acknowledgement}
This work is supported by the DoJ grants 15PJDP-21-GK-03269-MECP and 15PJDP-22-GK-03107-MECP. 
\end{acknowledgement}

%
%
\bibliographystyle{plain}
\bibliography{references}

%
\newpage
\section*{Appendix}
\addcontentsline{toc}{section}{Appendix}

\begin{table}[htbp]
\caption{Grooming Strategies as described in \cite{ring21}}
\label{app_tab:1}       
%
%
\begin{tabular}{p{0.3\textwidth}p{0.6\textwidth}}
\hline\noalign{\smallskip}
Strategy  & Description \\
\noalign{\smallskip}\hline\noalign{\smallskip}
Coercion & Peer-pressure, Implicit and overt threats, Guilt \\
Bragging &  Highlighting positive features around oneself and offers to teach sexual acts \cite{kingma2014} \\
Discuss Images &  Discussion, questions, and responses by one participant to the other related to images (both sexual and non-sexual) \\
Negative Physical &  Comments about having some negative element of physical appearance or body \\
Negative Life &  Stories or comments about not having a good family life or not getting along with family members \\
Teaching & Comments in which one participant offers to help the other participant learn something. The entry can be sexual or non-sexual \\
Personal Compliments & Providing positive comments about matureness, body, etc. of other \\
Reverse Power & False security meant to imply that the child is an equal/has control in the situation  \\
Sexual History & Comments about previous sexual experiences and whether or not the other participant enjoys sexual acts they have engaged in in the past \\
Willingness & Assessment of whether or not someone would do or try something; Questions related to sex acts which could be performed in person \\
Roleplay & Pretending to have sexual intercourse while actively online \\
Secrecy & Requests or statements expressing a need for discretion \\
\hline
\end{tabular}
\end{table}

\end{document}